\documentclass[accepted]{uai2021} 

\usepackage[american]{babel}

\usepackage{natbib} 
\usepackage{mathtools} 
\usepackage{booktabs} 
\usepackage{tikz} 

\usepackage{graphicx}
\usepackage[font=small,labelfont=bf]{caption}
\usepackage{subcaption}
\usepackage{amssymb}
\usepackage{amsmath}
\usepackage{amsfonts}
\usepackage[bottom]{footmisc}
\usepackage{wrapfig}
\usepackage[ruled]{algorithm}
\usepackage{algcompatible}
\usepackage{siunitx}
\usepackage{array}

\usepackage{tabularx}

\usepackage{multirow}

\usepackage{amsmath}
\usepackage{adjustbox}
\usepackage{bm}
\newcommand{\dataset}{\mathcal{D}}

\makeatletter
\newcommand{\algcolor}[2]{%
  \hskip-\ALG@thistlm\colorbox{#1}{\parbox{\dimexpr\linewidth-2\fboxsep}{\hskip\ALG@thistlm\relax #2}}%
}

\makeatother

\newcommand{\squishlist}{
   \begin{list}{$\bullet$}
    { \setlength{\itemsep}{2pt}    \setlength{\parsep}{0pt}
      \setlength{\topsep}{5pt}     \setlength{\partopsep}{0pt}
      \setlength{\leftmargin}{1.35em} \setlength{\labelwidth}{1em}
      \setlength{\labelsep}{0.5em} } }

\newcommand{\squishend}{
    \end{list}  }

\title{
Certification of Iterative Predictions in Bayesian Neural Networks
}

\author[1]{\href{mailto:Matthew Wicker <matthewwicker@cs.ox.ac.uk>?Subject=Your UAI 2021 paper}{\ \ Matthew Wicker* }{}}
\author[1]{\ \ \ \  Luca Laurenti*}
\author[1]{\ \ \ \ Andrea Patane}
\author[2]{\ \ \ \ Nicola Paoletti}
\author[1]{\ \ \ \ Alessandro Abate}
\author[1]{Marta Kwiatkowska}
\affil[1]{%
    Department of Computer Science\\
    University of Oxford\\
    Oxford, UK
}
\affil[2]{%
    Department of Computer Science \\
    Royal Holloway University of London\\
    London, UK
}

\begin{document}

\newtheorem{myexam}{Example}
\newtheorem{problem}{Problem}
\newtheorem{proposition}{Proposition}
\newtheorem{proof}{Proof}
\newtheorem{theorem}{Theorem}
\newtheorem{assumption}{Assumption}
\newtheorem{definition}{Definition}

\newcommand{\acmt}[1]{{\color{magenta}[AA: #1]}}
\newcommand{\ap}[1]{{\color{orange}[AP: #1]}}
\newcommand{\LL}[1]{{\color{red}[LL: #1]}}
\newcommand{\MRW}[1]{{\color{blue}[MW: #1]}}
\newcommand{\JC}[1]{{\color{cyan}[JC: #1]}}
\newcommand{\np}[1]{{\color{red!50!black}[NP: #1]}}
\newcommand{\mk}[1]{{\color{red}[MK: #1]}}

\newcommand{\jj}[1]{{\color{purple}[JJ: #1]}}
\newcommand\HS{\mathcal{H}}

\newcommand{\pX}{\mathbf{x}}

\maketitle
\begin{abstract}
We consider the problem of computing reach-avoid probabilities for iterative predictions made with Bayesian neural network (BNN) models.
Specifically, we leverage bound propagation techniques and backward recursion to compute lower bounds for the probability that trajectories of the BNN model reach a given set of states while avoiding a set of unsafe states. We use the lower bounds in the context of control and reinforcement learning to provide safety certification for given control policies, as well as to synthesize control policies that improve the certification bounds. On a set of benchmarks, we demonstrate that our framework can be employed to certify policies over BNNs predictions for problems of more than $10$ dimensions, and to effectively synthesize policies that significantly increase the lower bound on the satisfaction probability. 
\end{abstract}

\section{Introduction}


While retaining the main advantages intrinsic to deep learning, \textit{Bayesian neural networks} (BNNs)  reason about uncertainty in a principled and probabilistic manner, making them a particularly appealing model class for tackling \textit{safety-critical} scenarios.
In principle, their predictive uncertainty can be propagated through the decision pipeline to enable formal evaluation and analysis of a system under unknown conditions \citep{mcallister2016data}, which can model partial knowledge of the system as well as its intrinsic stochasticity \citep{depeweg2016learning}. 
%
%
%

In scenarios such as sequential planning, time-series forecasting/control and model-based reinforcement learning, to evaluate a model  w.r.t.\ a control policy (or strategy) one often needs to be able to make several predictions correlated across time \citep{liang2005bayesian}.
While multiple models can be learned for each time step, a common setting is for these predictions to be made iteratively by the same machine learning model \citep{huang2020deep}, 
where the predicted model state at each step is a function of the model configuration at the previous step and possibly an additional control input. 
We refer to this setting as \emph{iterative predictions}. 
The challenge with BNN models is that they output probability distributions, posing the problem of  successive predictions over a stochastic input.  
Even when the BNN posterior weights are estimated using analytical approximations, its deep and non-linear nature makes iterative predictions with BNNs an analytically intractable problem \citep{neal2012bayesian}. To
the best of our knowledge, computing formal bounds on the probability of BNN-based iterative predictions remains an open problem. 
Such bounds would enable one to provide safety guarantees over a given (or learned) control policy,  which is a necessary precondition before deploying the policy in a real-world environment \citep{polymenakos2020safety,vinogradska2016stability}. 
%

In this paper, we develop a method for the computation of probabilistic guarantees for iterative predictions with BNNs over \emph{reach-avoid} specifications. 
A reach-avoid specification, also known as constrained reachability \citep{SA13b}, requires that the trajectories of a dynamical system reach a goal region over a given (finite) time horizon, whilst avoiding a given set of unsafe states.
Probabilistic reach-avoid is a key property for formal analysis of stochastic processes \citep{abate2008probabilistic}, underpinning richer temporal logic specifications \citep{model_checking,a3c,cauchi2019efficiency}. 
%
Even though the exact computation of reach-avoid probabilities for iterative prediction with BNNs is analytically intractable, we show how to derive a guaranteed lower bound 
by solving a backward iterative problem obtained via a discretisation of the state space.  
In particular, starting from the final time step, we back-propagate lower bounds to reach-avoid probabilities through previous time steps and for each discretised portion of the state-space, beginning from the goal region.
The propagation of bounds through consecutive time steps leverages bound propagation techniques for BNNs~\citep{wicker2020probabilistic}.
%
By further combining these with bound propagation techniques for (non-Bayesian) neural networks (NNs) \citep{gowal2018effectiveness,gehr2018ai2}, we then discuss how the resulting lower bound can be employed to provide certificates for NN policies learned over the BNN dynamical system.
Finally, we demonstrate how our bound can be used to tackle the synthesis problem, where given an initial policy we seek to maximise the lower bound associated to a given reach-avoid specification.  
 
In a set of case studies, we confirm the  scalability of our methodology. 
We begin by considering four planar control problems involving obstacle layouts of varying complexity.
We then study the scalability of our framework on two locomotion problems from the Mujoco robotic physics simulator \citep{ray2019benchmarking}.
With our  approach,  we can derive probabilistic reach-avoid certifications for planar control tasks, including a 25-dimensional car agent.
Finally, we demonstrate how controllers can be successfully improved by using our synthesis algorithm. 
In summary, this paper makes the following contributions: 
\begin{itemize}
    \item We show how probabilistic reach-avoid for iterative prediction with BNNs can be formulated as the solution of a backward computation problem, and design an algorithm for the lower bounding of the latter. 
    \item We discuss how our lower-bound can be used for policy certification and, further, for synthesising NN control policies via dynamic programming.
    \item We demonstrate the applicability of our methodology on a set of case studies with more than 10 dimensions.
\end{itemize}

\section{Related Work}
Certification of machine learning models is a rapidly growing area~\citep{gehr2018ai2,katz2017reluplex,gowal2018effectiveness}. 
While most of these methods have been designed for deterministic NNs,
recently safety analysis of Bayesian machine learning models has been studied both for Gaussian processes (GPs)~\citep{grosse2017wrong,cardelli2018robustness,blaas2020adversarial} and BNNs~\citep{athalye2018obfuscated,cardelli2019statistical, wicker2020probabilistic},  including methods for adversarial training  \citep{liu2018adv,wicker2021bayesian}. 
The above works, however, focus exclusively on the input-output behaviour of the  models, that is, can only reason about static properties. 
Conversely, the problem we tackle in this work has additional complexity, as we aim to formally reason about iterative predictions, i.e., trajectory-level behaviour of a BNN interacting in closed-loop with a controller.
Iterative predictions have been widely studied for Gaussian processes  \citep{girard2003gaussian} and safety guarantees have been proposed in this setting in the context of model-based RL with GPs \citep{jackson2020safety,polymenakos2019safe,berkQuad}. 
However, all these works are specific to GPs and cannot be extended to BNNs, whose posterior predictive distribution is intractable and non-Gaussian even for the more commonly employed approximate Bayesian inference methods \citep{neal2012bayesian}.

Various recent works consider verification or synthesis of RL schemes against reachability specifications 
\citep{sun2019formal,konighofer2020safe,bacci2020probabilistic}.
None of these approaches, however, support both continuous state-action spaces and probabilistic models, as in this work. 
Continuous action spaces are supported in \citep{hasanbeig2020certified}, where the authors provide RL schemes for the synthesis of policies maximising given temporal requirements. However, the guarantees resulting from these model-free algorithms are asymptotic, and thus of different nature than those in this work.  
The work of \cite{HVA17} integrates Bayesian inference and formal verification over control models, additionally proposing strategy synthesis approaches for active learning~\citep{cHVA16,bcWA19}. In contrast to our paper these works do not support unknown noisy models learned via BNNs.

\section{Background}
\label{sec:backgoundNN}
In this section we briefly review BNNs and modeling of discrete-time dynamical systems with BNNs.
\paragraph{Bayesian Neural Networks} 

Let $f^w:\mathbb{R}^{m}\to\mathbb{R}^n$ be a feed-forward NN architecture, where $w\in \mathbb{R}^{n_w}$ is the vector containing all the weights and biases of the network. 
BNNs extend NNs by having a prior distribution placed over the network parameters, $p_{\mathbf{w}}(w)$, with $\mathbf{w}$ being the vector of random variables associated to the weights vector. 
Given a dataset $ \dataset$, a BNN posterior, $p_{\mathbf{w} } (w \vert \dataset)$,  is inferred approximately by means of Bayes' rule~\citep{neal2012bayesian}. 
Unfortunately, $p_{\mathbf{w} } (w \vert \dataset)$ is analytically intractable. 
Thus, various techniques have been developed to approximate $p_{\mathbf{w} } (w \vert \dataset)$, including Hamiltonian Monte Carlo (HMC)~\citep{neal2012bayesian} and Variational Inference (VI) \citep{blundell2015weight}.
While we conduct experiments on VI, the techniques we describe are general and can be employed to HMC methods, e.g., by using the approach of \cite{wicker2021bayesian}. 


\paragraph{Iterative Predictions of BNNs}
Given a trained BNN, $f^{\mathbf{w}}$, we consider its associated dynamical system described by the following discrete-time stochastic control process:
\begin{align}
\label{Eqn:SystemEqn}
  & \mathbf{x}_k = f^{\mathbf{w}}(\mathbf{x}_{k-1},\mathbf{u}_{k-1})+\mathbf{v}_k, \quad  \mathbf{u}_k = \pi_k(\mathbf{x}_k), \\ & k\in \mathbb{N}_{>0},\,\mathbf{x}_{k},\mathbf{v}_k\in \mathbb{R}^n,\, \mathbf{u}_k \in \mathcal{U}\subseteq \mathbb{R}^{c}, 
   \nonumber
\end{align}
where $\mathbf{v}_k$ is a random variable modelling an  additive noise term with stationary, zero-mean Gaussian distribution $\mathcal{N}(\bar{x}|0,\sigma^2\cdot I),$ where $I$ is the identity matrix. 
The vector $\mathbf{x}_k$ is the model state at time $k$; $\mathbf{u}_k$ represents the control input applied at time $k$, selected from an admissible, compact set $\mathcal{U}\subset \mathbb{R}^c$ by 
a (deterministic) feedback Markov strategy (policy) $\pi:\mathbb{R}^{n}\times \mathbb{N} \to \mathcal{U}$.\footnote{For our settings, optimal strategies are time-dependent and Markov~\citep{bertsekas2004stochastic}.}  Intuitively, the model in Eqn \eqref{Eqn:SystemEqn} represents a noisy controlled discrete-time stochastic process whose time evolution is given by iterative predictions of the BNN $f^{\mathbf{w}}$, and is controlled by $\pi$.
In this setting, $f^{\mathbf{w}}$ defines the transition probabilities of the system and $p_{\mathbf{w}}(w|\mathcal{D})$ is employed to estimate the posterior predictive distribution $p(\bar{x}|(x,u), \mathcal{D})$ that describes the probability density of the system at the next time step being $\bar{x}$, given that the current state and action are $(x,u)$, and it is defined as:
\begin{align*}
    p(\bar{x}|(x,u), \mathcal{D})=\int_{\mathbb{R}^{n_w}} \mathcal{N}(\bar{x}|f^w(x,u),\sigma^2\cdot I)p_{\mathbf{w}}(w| \mathcal{D}) dw,
\end{align*}
where $\mathcal{N}(\bar{x}|f^w(x,u),\sigma^2\cdot I)$ is the Gaussian likelihood induced by $\mathbf{v}_k$ 
and centered at the NN output \citep{neal2012bayesian}. 
Observe that the posterior predictive distribution induces a probability density function over the state space.
In iterative prediction settings this implies that at each step the state vector $\mathbf{x}_k$ fed into the BNN is a random variable.
Hence, a principled propagation of the BNN uncertainty through consecutive time steps poses the problem of predictions over stochastic inputs.
In Section \ref{sec:ReachAvoid} we will tackle this for the particular case of reach-avoid properties, by designing a backward computation scheme that starts its calculations from the goal region.
We remark that $p(\bar{x}|(x,u), \mathcal{D})$ is defined by marginalizing over $p_{\mathbf{w}}(w|\mathcal{D})$.
Hence, the particular $p(\bar{x}|(x,u), \mathcal{D})$ depends on the specific approximate inference method employed to estimate the posterior distribution.
As such, the bounding results that we derive are valid only for a specifically trained BNN.

\section{Problem Formulation}
\label{sec:ProbForm}

For an action $u\in \mathbb{R}^c$, a subset of states $X\subseteq \mathbb{R}^m$ and a starting state $x\in \mathbb{R}^m$, 
we call $T(X|x,u)$ the \emph{stochastic kernel} associated to the dynamical system. 
$T(X|x,u)$  describes the one-step transition probability of the model of Eqn.~\eqref{Eqn:SystemEqn} and is defined by integrating the predictive posterior distribution with input $(x,u)$ over $X$:
\begin{align*}
    T(X|x,u)=& \int_X p(\bar{x}|(x,u),\mathcal{D}) d \bar{x}. 
\end{align*}  
Note that the integral is defined here over the state space ($\mathbb{R}^n$).
In what follows, it will be convenient at times to work in the parameter space of the BNN instead. To do so, we can re-write the stochastic kernel by applying Fubini's theorem to switch the integration order, thus obtaining: 
\begin{align*} 
  &T(X|x,u)=\\
  &\quad \quad\int_{\mathbb{R}^{n_w}} \left[ \int_X \mathcal{N}(\bar{x}|f^w(x,u),\sigma^2\cdot I) d\bar{x} \right] p_{\mathbf{w}}(w|\mathcal{D}) dw.
\end{align*}
From the definition of $T$ it follows that,  under a strategy $\pi$ and for a given initial condition $x_0$, $\mathbf{x}_k$  is a Markov process with a well defined probability measure $\Pr$ uniquely generated by the stochastic kernel $T$ \citep[Proposition 7.45]{bertsekas2004stochastic} and such that for $X_0,X_k \subseteq \mathbb{R}^n$:
\begin{align*}
& \Pr[\pX_0\in X_0] =  \mathbf{1}_{X_0}(x_0), \\
& \Pr[\pX_k\in X_k  |\pX_{k-1}=x, \pi] =  T^{}(X_k|x,
\pi_{k-1}(x)).
\end{align*}
The definition of $\Pr$ allows one to make probabilistic statements over the model in Eqn \eqref{Eqn:SystemEqn}. 
In Problem \ref{Prob:verification} we consider  probabilistic reach-avoid, that is the probability that a trajectory of $\mathbf{x}_k$ reaches a goal region within the state space, whilst always avoiding a given set of (bad) states. 

\begin{problem}[Computation of Probab.\ Reach-Avoid]
\label{Prob:verification}
Given a strategy $\pi$, a goal region $\mathrm{G}\subseteq \mathbb{R}^m$, a finite-time horizon $[0,N]\subseteq \mathbb{N},$  and a safe set $\mathrm{S}\subseteq \mathbb{R}^m$ such that $\mathrm{G}\cap \mathrm{S}= \emptyset $, compute for any given $x_0 \in \mathrm{G}\cup \mathrm{S}$
\begin{align}
\label{Eqn:ProbForm}
   P_{reach}&(\mathrm{G},\mathrm{S},x_0,[0,N]|\pi)= 
  Pr\big[\exists k \in [0,N], \pX_k \in \mathrm{G}\, \wedge \\
\nonumber   &\hspace{2cm}\forall 0 \leq k' < k, \pX_{k'} \in \mathrm{S} \mid \pX_0=x_0,\pi\big].
\end{align}
\end{problem}
Note that, in Problem \ref{Prob:verification}, the strategy $\pi$ is given, and the goal is to quantify the probability with which the trajectories of $\mathbf{x}_k$ satisfy the given specification. 
%

In Problem~\ref{Prob:Syntesis} below we generalise the previous problem and seek to synthesise a controller $\pi$ that guarantees that $P_{reach}(\mathrm{G},\mathrm{S},x_0,[0.N]|\pi)$ is above a given threshold $\delta$.  

\begin{problem}[Strategy Synthesis for Probab.\ Reach-Avoid] 
\label{Prob:Syntesis}
For a given tolerance $0<\delta<1$ and $x_0 \in \mathrm{G}\cup \mathrm{S}$, find a strategy $\pi:\mathbb{R}^{ n}\times \mathbb{R}_{ \geq 0}\to\mathbb{R}^c$ such that
\begin{align}
  P_{reach}(\mathrm{G},\mathrm{S},x_0,[0,N]\mid \pi)>1-\delta.
\end{align}
\end{problem}


\textbf{Outline of the Approach } 
In Section \ref{sec:ReachAvoid} we show how $P_{reach}(\mathrm{G},\mathrm{S},x,[k,N]|\pi)$ can be formulated as the solution of a backward iterative computational procedure, where the uncertainty of the BNN is propagated backward over time starting from the goal region. We will show that such a formulation of $P_{reach}$ has two main advantages.
Firstly, it allows us to define techniques for certification of BNNs to compute a sound lower bound on  $P_{reach}$, thus guaranteeing that the process $\pX_k$ satisfies the specification with a given probability (Section \ref{sec:LowerBound}). 
Secondly, relying on the differentiability of the resulting lower bound, it allows one to synthesize control strategies to improve the lower bound on the reach-avoid probability.
\section{Probabilistic Reach-Avoid}
\label{sec:ReachAvoid}

In this section we show how $P_{reach}(\mathrm{G},\mathrm{S},x,[k,N]|\pi)$ can be formulated as the solution of a backward iterative procedure, which will allow us to compute a lower bound on its value. 

Given a time $0\leq k<N$ and  strategy $\pi,$ consider the value functions $V_k^{\pi}:\mathbb{R}^n\to [0,1]$, recursively defined as  
\begin{align}
    &\hspace*{-0.1cm}\nonumber V_N^{\pi}(x)=\mathbf{1}_\mathrm{G}(x), \\
    &\hspace*{-0.1cm}V_k^{\pi}(x)=\mathbf{1}_\mathrm{G}(x)+\mathbf{1}_\mathrm{S}(x){\hspace*{-0.1cm} \int \hspace*{-0.1cm} V^{\pi}_{k+1}(\bar{x}) p\big(\bar{x} | (x, \pi_{k}(x)),\mathcal{D}\big) d\bar{x}}. 
    \label{Eqn:ExactValueFunc}
\end{align}
Intuitively, $V_k^{\pi}$ is computed backwards starting from the goal region $\mathrm{G}$ at $k=N$, where it is initialised at value $1$.
The computation then proceeds backwards for each state $x$ by combining the current values with the transition probabilities coming from Eqn~\eqref{Eqn:SystemEqn}. 
The following proposition, proved in the Supplementary Material, guarantees that $ V_0^{\pi}(x)$ is indeed a characterisation of $P_{reach}(\mathrm{G},\mathrm{S},x,[0,N]|\pi)$.
\begin{proposition}
\label{th:CorrectnessBack}
For $0\leq k\leq N$ and $x_0 \in \mathrm{G}\cup\mathrm{S},$ it holds that 
$$  
P_{reach}(\mathrm{G},\mathrm{S},x_0,[k,N]|\pi)= V_k^{\pi}(x).
$$
\end{proposition}
\noindent
The backward recursion in Eqn~\eqref{Eqn:ExactValueFunc} does not generally admit a solution in closed-form, as it would require integrating over the BNN posterior predictive distribution, which is in general analytically intractable.
A computational scheme for lower bounding this quantity is derived in the following section. 
%
\subsection{Lower Bound on $P_{reach}$}\label{sec:LowerBound}
We develop a computational approach based on the discretisation of the state space,  and on the backward formulation of Eqn \eqref{Eqn:ExactValueFunc}, for calculating a lower bound for $P_{reach}$. 
As this represents a conservative estimation, the lower bound on reach-avoid can thus be used to provide probabilistic certification of a strategy controlling the BNN dynamical system.  
\begin{figure}[h]
    \centering
    \includegraphics[width=0.5\textwidth]{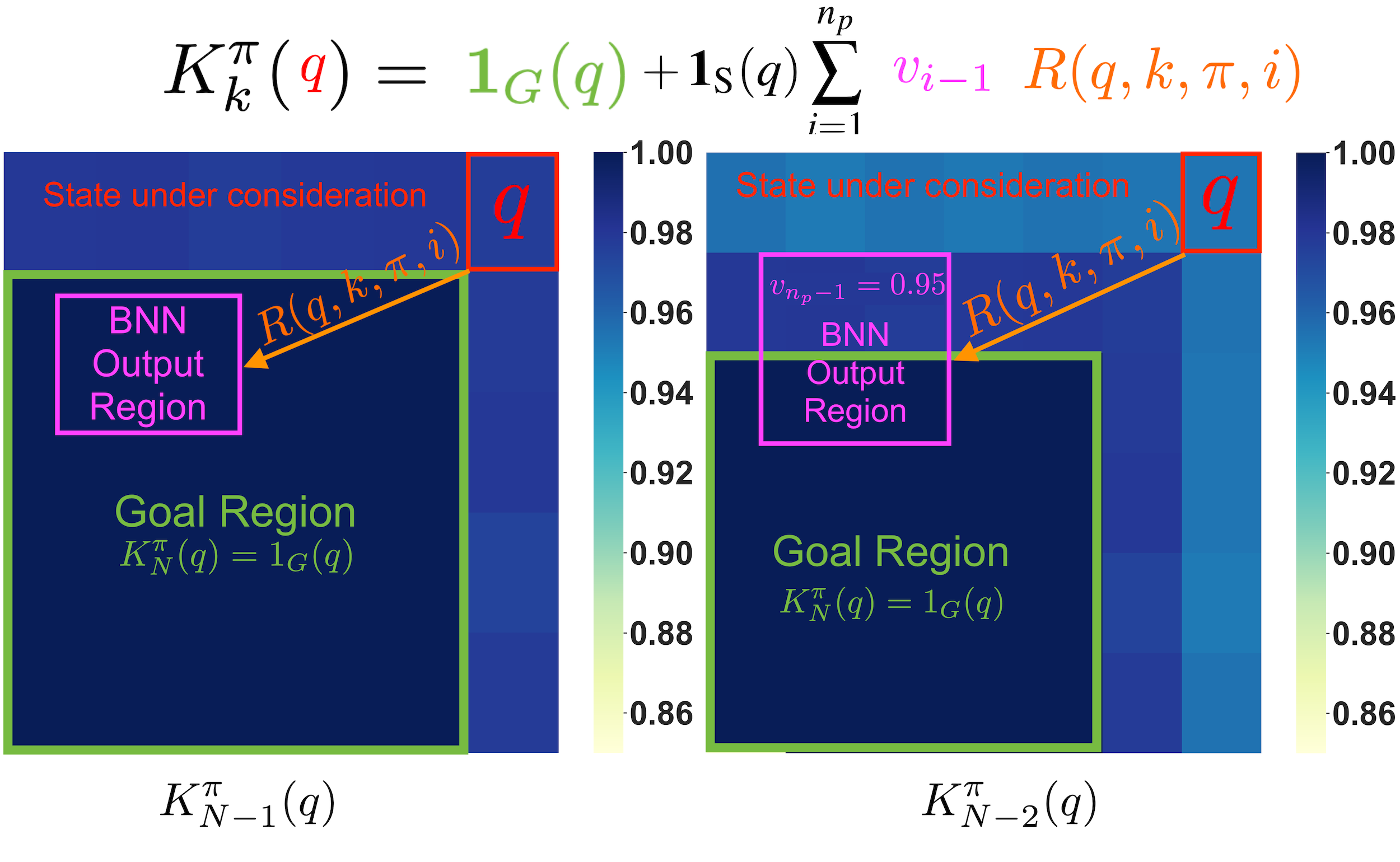}
    \caption{
    Examples of functions $K^{\pi}_{k}$, which are lower bounds of $V^{\pi}_k$ for any $0\leq k\leq N$. 
     On the left, we consider the first step of our backward algorithm, where we compute $K^{\pi}_{N-1}(q)$ by computing the probability that $\mathbf{x}_{N}\in \mathrm{G}$ given that $\mathbf{x}_{N-1}\in q.$  On the right, we consider the subsequent step. We outline the state we want to verify in red and the goal region in green. With the orange arrow we represent the 0.95 transition probability of the BNN dynamical model, and in pink we represent the worst-case probabilities spanned by the BNN output. On top, we show where each of these key terms comes into play in Eqn~\eqref{Eqn:ValueFuncDiscre}.  }\label{fig:notation_visualization}
\end{figure}
%
%

The proposed computational approach is illustrated in Figure \ref{fig:notation_visualization}. 
Let $Q=\{q_1,...,q_{n_q} \}$ be a partition of $\mathrm{S} \cup \mathrm{G}$ in $n_q$ regions. 
We denote with  $z:\mathbb{R}^n\to Q$ the function that associates to a state in $\mathbb{R}^n$ the corresponding partitioned state in $Q$. 
For each $0\leq k \leq N$ we iteratively build a  set of functions $K^{\pi}_{k}: Q \rightarrow [0,1]$ such that for all $x\in \mathrm{G}\cup \mathrm{S}$ we have that  $K^{\pi}_{k}(z(x))\leq V^{\pi}_{k}(x)$.
Intuitively, $K^{\pi}_{k}$ provides a discretised lower bound for the value functions on the computation of $P_{reach}$.

The functions $K^{\pi}_{k}$ are obtained by propagating backward the BNN predictions from time $k=N$, where we set $K^{\pi}_{N}(q)=\mathbf{1}_{\mathrm{G}}(q)$, with $\mathbf{1}_{\mathrm{G}}(q)$ being the indicator function (that is, $1$ if $q\subseteq \mathrm{G}$ and $0$ otherwise).  Then, for each $k<N$, we first discretize the set of possible probabilities in $n_p$ sub-intervals $0=v_0\leq v_1\leq...\leq v_{n_p}=1$.
Hence, for any $q\in Q$ and probability interval $[v_i,v_{i+1}]$, we compute a lower bound $R(q,k,\pi,i)$ on the probability that, starting from any state in $q$ at time $k$,  we reach in the next step a region that has probability $\in [v_i,v_{i+1}]$ of safely converging to the goal region.
The resulting values are used to build $K^{\pi}_{k}$ (as we will detail in Eqn \eqref{Eqn:ValueFuncDiscre}). For a given $q\subset \mathrm{S}$, $K^{\pi}_{k}(q)$ is obtained as the sum over $i$ of $R(q,k,\pi,i)$ multiplied by $v_{i-1}$, i.e.,  the lower value that $K^{\pi}_{k+1}$ obtains in all the states of the $i-th$ region. Note that the discretisation of the probability values does not have to be uniform, but can be adaptive for each $q\in Q$. A heuristic for picking the value of thresholds $v_i$ will be given in Algorithm \ref{alg:backwardsreachability}. 
In what follows, we formalise the intuition behind this computational procedure.

\paragraph{Lower Bounding of the Value Functions} 
For a given strategy $\pi$, consider  a constant $\eta\in (0,1)$ and $\epsilon=\sqrt{2\sigma^2}\text{erf}^{-1}(\eta)$, which are used to bound the value of the noise, $\mathbf{v}_k$, at any given time.\footnote{The thresholds are such that it holds that $Pr(|\mathbf{v}_k^{(i)}|\leq \epsilon)=\eta$. In the experiments of Section \ref{sec:experiments} we select $\eta=0.99$.} 
Then, for $0\leq k<N$, consider the functions $K^{\pi}_k:Q \to [0,1]$ defined recursively as follows:  
\begin{align}
    &K^{\pi}_N(q)=\mathbf{1}_{\mathrm{G}}(q), \label{Eqn:ValueFuncDiscreTrivial}\\
    &K^{\pi}_k(q)=\mathbf{1}_{\mathrm{G}}(q) + \mathbf{1}_{\mathrm{S}}(q) \sum_{i=1}^{n_p}v_{i-1}R(q,k,\pi,i)  ,%
    \label{Eqn:ValueFuncDiscre}
\end{align}
where
\begin{align}
&R(q,k,\pi,i)= \eta^n \int_{H^{q,\pi,\epsilon}_{k,i}}p_{\mathbf{w}}(w|\mathcal{D})   dw, \label{eq:def_of_R}\\ 
&  H^{q,\pi,\epsilon}_{k,i}=\{ w \in \mathbb{R}^{n_w} | \, \forall x\in q, \forall \gamma \in [-\epsilon,\epsilon]^n, \text{ it holds that: } \nonumber \\
  &\hspace{0.3cm}  v_{i-1} \leq  K_{k+1}^{\pi}(q') \leq v_{i}, \text{ with $q'=z(f^w(x,\pi_{k}(x))+\gamma)$} \}. \nonumber
  \end{align}
The key component for combining the above computations together is
$R(q,k,\pi,i) $, which bounds the probability that, starting from $q$ at time $k$, we have that $\mathbf{x}_{k+1}$ will be in a region $q'$ such that $K_{k+1}^{\pi}(q')\in [v_i,v_{i+1}]$.
In fact, $H^{q,\pi,\epsilon}_{k,i}$ defines the weights for which that is true, so that integration of the posterior $p_{\mathbf{w}}(w|\mathcal{D})$ over the  $H^{q,\pi,\epsilon}_{k,i}$ will return the probability mass for the BNN dynamical system transitioning from $q$ to $q'$ with probability in $[v_i,v_{i+1}]$.
The computation of Eqn \eqref{Eqn:ValueFuncDiscre} then reduces to computing the set of weights $H^{q,\pi,\epsilon}_{k,i}$, which we call the \textit{projecting weight set}.
A method to compute a safe under-approximation $ \bar{H}\subseteq  H^{q,\pi,\epsilon}_{k,i}$ is discussed below. 
Before describing that, we analyze the correctness of the above recursion. 
\begin{theorem}
\label{th:VerificationLoerBound}
Given $x\in \mathbb{R}^n$, for any  $k\in \{0,...,N\}$ and $q = z(x)$,  assume that $H^{q,\pi,\epsilon}_{k,i} \cap H^{q,\pi,\epsilon}_{k,j}=\emptyset$ for $i\neq j.$ Then:
$$ \inf_{x\in q} V_{k}^{\pi}(x)\geq  K^{\pi}_k(q).$$
\end{theorem}
A proof of Theorem \ref{th:VerificationLoerBound} is given in the Supplementary Material.
Note that the assumption on the null intersection between different projecting weight sets required in Theorem \ref{th:VerificationLoerBound} can always be enforced by taking their intersection and complement.

\paragraph{Computation of Projecting Weight Sets}
\label{sec:WeightSetsComputation}
Theorem \ref{th:VerificationLoerBound} allows us to compute a safe lower bound to Problem \ref{Prob:verification}, by relying on an abstraction of the state space, that is, through the computation of $K^{\pi}_{0}(q)$.
This can be evaluated once the projecting set of weight values $H^{q,\pi,\epsilon}_{k,i}$ associated to $[v_{i-1},v_i]$ is known.\footnote{In the case of Gaussian VI the integral of Equation \eqref{eq:def_of_R} can be computed in terms of the $\mathit{erf}$ function, whereas more generally Monte Carlo or numerical integration techniques can be used.}
Unfortunately, direct computation of $H^{q,\pi,\epsilon}_{k,i}$ is intractable. 
Nevertheless, a method for its lower bounding was developed by \cite{wicker2020probabilistic} in the context of adversarial perturbations for one-step BNN predictions, and can be directly adapted to our settings.
%

The idea is that a safe approximation $\bar{H} \subseteq H^{q,\pi,\epsilon}_{k,i}$ is built by sampling weight boxes of the shape $\hat{H} = [w^L,w^U]$, according to the posterior, and checking whether: 
\begin{align}
    v_{i-1}\leq K_{k+1}^\pi (z(f^w({x},\pi_{k}(x)  ) +  \gamma ) )\leq v_{i},  \nonumber \\  \forall x \in q, \, \forall w \in \hat{H}, \, \forall \gamma \in [-\epsilon,\epsilon]^n. \label{eq:one_step_condition}
\end{align}
Finally, $\bar{H}$ is built as a disjoint union of boxes $\hat{H}$  satisfying the above condition. 
%
In order to apply this method to our setting, we propagate the abstract state $q$ through the policy function $\pi_{k}(x)$, so as to obtain a bounding box $\widehat{\Pi} = [\pi^L,\pi^U]$ such that $\pi^L \leq \pi_{k}(x) \leq \pi^U$ for all $x \in q$.
In the experiments we focus on the case in which $\pi_{k}(x)$ is given by a NN controller, so that methods for bound propagation of NNs can be used for the computation of $\widehat{\Pi}$ \citep{gowal2018effectiveness,gehr2018ai2}.
%
%
%
The results from \cite{wicker2020probabilistic} can then be used to propagate $q$, $\widehat{\Pi}$ and $\hat{H}$ through the BNN, that is, to compute values $f^{L}_{q,\epsilon,k}$ and $f^{U}_{q,\epsilon,k}$ such that, for all  $x \in q, \gamma \in [-\epsilon,\epsilon]^n, w \in \hat{H}$, it holds that:
\begin{align}\label{eq:ibp_final_results}
  f^{L}_{q,\epsilon,k} \leq f^w({x},\pi_{k}(x)  ) +  \gamma  \leq f^{U}_{q,\epsilon,k}.
\end{align}
Furthermore, $f^{L}_{q,\epsilon,k}$ and $f^{U}_{q,\epsilon,k}$ are differentiable w.r.t.\ to the input vector.
Finally, the two bounding values can be used to check whether or not the condition in Eqn \eqref{eq:one_step_condition} is satisfied, by simply checking whether $ [   f^{L}_{q,\epsilon,k} ,  f^{U}_{q,\epsilon,k} ] $  propagated through $K_{k+1}^{\pi}$ is within $[v_i,v_{i+1}]$. Now that we have the necessary ingredients, in the following we describe our algorithm for the lower bounding of $P_{reach}$.



\textbf{Probabilistic Reach-Avoid Algorithm }
In Algorithm~\ref{alg:backwardsreachability} we summarize our approach for computing a lower bound for Problem \ref{Prob:verification}. 
For simplicity of presentation, we consider the case $n_p=2$, (i.e., we partition the range of probabilities in just two intervals $[0,v_1],$ $[v_1,1]$ - the case $n_p>2$ follows similarly). 
The algorithm proceeds by first initializing the reach-avoid probability for the partitioned states $q$ inside the goal region $G$ to $1$, as per Eqn \eqref{Eqn:ValueFuncDiscreTrivial}.
Then, for each of the $N$ time steps and for each one of the remaining partition states $q$, in line $4$ we set the threshold probability $v_1$ equal to the maximum value that $K^{\pi}$ attains at the next time step over the states in the neighbourhood of $q$ (which we capture with a hyper-parameter $\rho_x > 0$). 
We found this heuristic for the choice of $v_1$ to work well in practice (notice that the obtained bound is formal irrespective of the choice of $v_1$, and different choices could potentially be explored). 
We then proceed in the computation of Eqn \eqref{Eqn:ValueFuncDiscre}. 
This computation is performed in lines $5$--$14$.
First, we initialise to the null set the current under-approximation of the projecting weight set, $\bar{H}$.
We then sample $n_s$ weights boxes $\hat{H}$ by sampling weights from the posterior, and expanding them with a margin $\rho_w$ heuristically selected (lines 6-8).
Then, for each of these sets we first propagate the state $q$, policy function, and weight set $\bar{H}$ to build a box $\bar{X}$ according to Eqn \eqref{eq:ibp_final_results} (line 9), which is then accepted or rejected based on the value that $K^{\pi}$ at the next time step attains in states in $\bar{X}$ (lines 10-12). $K^{\pi}_{N-i}(q)$ is then computed in line 14 by integrating $p_{\mathbf{w}}(w|\mathcal{D})$ over the union of the accepted sets of weights.

\begin{algorithm} 
\caption{Probabilistic Reach-Avoid for BNNs}\label{alg:backwardsreachability}\small
\textbf{Input:} BNN model $f^{\mathbf{w}}$, safe region $\mathrm{S}$, goal region $\mathrm{G}$, discretization  $Q$ of $\mathrm{S} \cup \mathrm{G}$, time horizon $N$, neural controller $\pi$, number of BNN samples $n_s$, weight margin $\rho_w$, state space margin $\rho_x$ \\
\textbf{Output:} Lower bound on $K^{\pi}$ \\
\vspace*{-0.35cm}
\begin{algorithmic}[1]
\STATE For all $0\leq k \leq N$ set $K^\pi_k(q)=1$ iff $q\subseteq \mathrm{G}$ and $0$ otherwise
\FOR {$k\gets N$ to $1$} 
    \FOR{$q \in Q \setminus \mathrm{G}$}
        \STATE $v_1 \gets \max_{x\in [q-\rho_x,q+\rho_x]} K^{\pi}_{k+1}(z(x)) $
        \STATE $\bar{H} \gets \emptyset$ \COMMENT{$\bar{H}$ is the set of safe weights}
        \FOR{desired number of samples, $n_s$}
            \STATE $w' \sim P(w | \mathcal{D})$
            \STATE $\hat{H} \gets [w'-\rho_w, w'+\rho_w]$
            \STATE \# Propagation according to $(\text{Eqn \eqref{eq:ibp_final_results})}$
            \STATE $\bar{X} := [ f^{L}_{q,\epsilon,k}, f^{U}_{q,\epsilon,k}] \gets \text{Prop.}( q,\,\pi,\,\hat{H},\, \gamma)$
            \IF{$K^{\pi}_{k+1}(\bar{X})\geq v_1$}
                \STATE $ \bar{H} \gets \bar{H} \bigcup \hat{H}$
            \ENDIF
        \ENDFOR
        \STATE $K^\pi_{k}(q) = v_1\cdot \eta^n\int_{\bar{H}}p_{\mathbf{w}}(w|\mathcal{D})dw$ \; (Eqn \eqref{Eqn:ValueFuncDiscre})
    \ENDFOR
\ENDFOR
\end{algorithmic}
\end{algorithm}
\section{Strategy Synthesis}
\label{Sec:Syntesis}
We now focus on the synthesis problem. More specifically, instead of bounding the reach-avoid probability for a given strategy $\pi$, we are interested in synthesising such a strategy.
In particular, we do this by finding the strategy $\pi^*$ that maximises the lower bound to $P_{reach}$ that we developed in the previous section.
Notice that, while no global optimality claim can be made about the strategy that we obtain, the maximisation of a lower bound guarantees that the true reach-avoid probability will still be greater than the improved bound obtained after the maximisation.
\begin{definition}\label{def:max-cert}
A strategy $\pi^*$ is called maximal certified (max-cert), w.r.t.\ to the discretised value function $K^\pi$,  if and only if, for all $x \in  \mathrm{G}\cup\mathrm{S}$, it satisfies  
$$  
K_{0}^{\pi^*}(z(x))=\sup_{\pi} K_{0}^{\pi}(z(x)),
$$
that is, the strategy $\pi^*$ maximises the lower bound of $P_{reach}$.
\end{definition}
\noindent
It follows that, if $K_{0}^{\pi^*}(z(x))>1-\delta$ for all $x \in G \cup S$, then the max-cert strategy $\pi^*$ is a solution of Problem 2. 
Note that a max-cert strategy is guaranteed to exist when the set of admissible controls $\mathcal{U}$ is compact \cite[Lemma 3.1]{bertsekas2004stochastic} (as we assume in this work). 
In the next theorem we show that a max-cert strategy can be computed via dynamic programming with a backward recursion similar to that of Eqn \eqref{Eqn:ValueFuncDiscre}.
\begin{theorem}
\label{Th:Syntesis}
For  $0\leq k<N$ and $0=v_0<...<v_{n_p}=1,$ define the functions $K^*_{k}:\mathbb{R}^n\to [0,1]$  recursively as follows
\begin{align*}
    &K^*_{k}(q)=\sup_{u\in \mathcal{U}} \big( \mathbf{1}_\mathrm{G}(q) + \mathbf{1}_\mathrm{S}(q) \sum_{i=1}^{n_p}v_{i}R(q,k,u,i)\big), 
\end{align*}
where $R(q,k,u,i)$ and $H^{q,u,\epsilon}_{k,i}$ are defined as in Eqn \eqref{eq:def_of_R}.
If $\pi^*$ is s.t.\ $K^*_0=K^{\pi^*}_0$, then $\pi^*$ is a max-cert strategy. 
Furthermore, for any $x$, it holds that $K^{\pi^*}_0(z(x))\leq P_{reach}( \mathrm{G},\mathrm{S},[0,N],x | \pi^*)$. 
\end{theorem}
A proof for Theorem \ref{Th:Syntesis} can be derived similarly as in  \cite[Theorem 2]{abate2008probabilistic}. 
Theorem \ref{Th:Syntesis} allows one to recursively compute a max-cert strategy, by selecting at each time step the action that maximizes the function $K$.
Note that the resulting $\pi^*$ will generally depend on the time step $k$.
We remark that Theorem \ref{Th:Syntesis} does not make any assumption on the form of $\pi$, so that any can be employed, as long as the set in which $\pi$ varies is parametrised by a compact set.
In the following we focus, in particular, on the case in which the set of allowed strategies $\mathcal{U}$ is parametrised by a NN controller $\pi$, which is of particular relevance for RL applications \citep{arulkumaran2017deep}. 
Namely, we show how a neural controller $\pi$ that builds on the lower bound can be computed using standard training methods for NNs.
\paragraph{Training of Certified NN Strategies } 
In Theorem \ref{Th:Syntesis}, the only term that depends on the input $\pi$ is
$\sum_{i=1}^{n_p}v_{i}R(q,k,\pi,i)$. 
Hence, in order to synthesise a strategy one needs to find the neural controller input, over $\mathcal{U}$, that maximizes the integral of $p_{\mathbf{w}}(w|\mathcal{D})$ over the projecting weight sets $H^{q,\pi,\epsilon}_{k,i}.$

Let $\mathcal{L}_{reward}$ be a (differentiable) reward function for the control problem at hand (which we obtain at training time by employing standard model-based RL algorithms \citep{arulkumaran2017deep}).
Our goal is to synthesize the parameters of $\pi_k$ such that $\mathcal{L}_{reward}$ is maximised, while also maximising the lower-bound to $P_{reach}$.
In order to do so, we proceed in a similar fashion to methods for adversarial training of NNs with bound propagation techniques  \citep{gowal2018effectiveness}.
Consider $P_{reach}^{LB}(\pi)$ to be the lower bound to probabilistic reach-avoid that we have developed in Section \ref{sec:ReachAvoid}.
Interestingly, because of differentiability of $P_{reach}^{LB}(\pi)$ the policy parameters can be optimised using standard out-of-the-box gradient descent methods for NNs. 
We remark that, even though Theorem \ref{Th:Syntesis} guarantees existence of a max-cert strategy, performing gradient descent does not guarantee to find one. However, it does provide significant local improvements of the reach-avoid probability around the  starting policy $\pi$, as we show in the next section.

\section{Experiments}
\label{sec:experiments}

\begin{figure*}[h]
    \centering
    \includegraphics[width=0.875\textwidth]{ 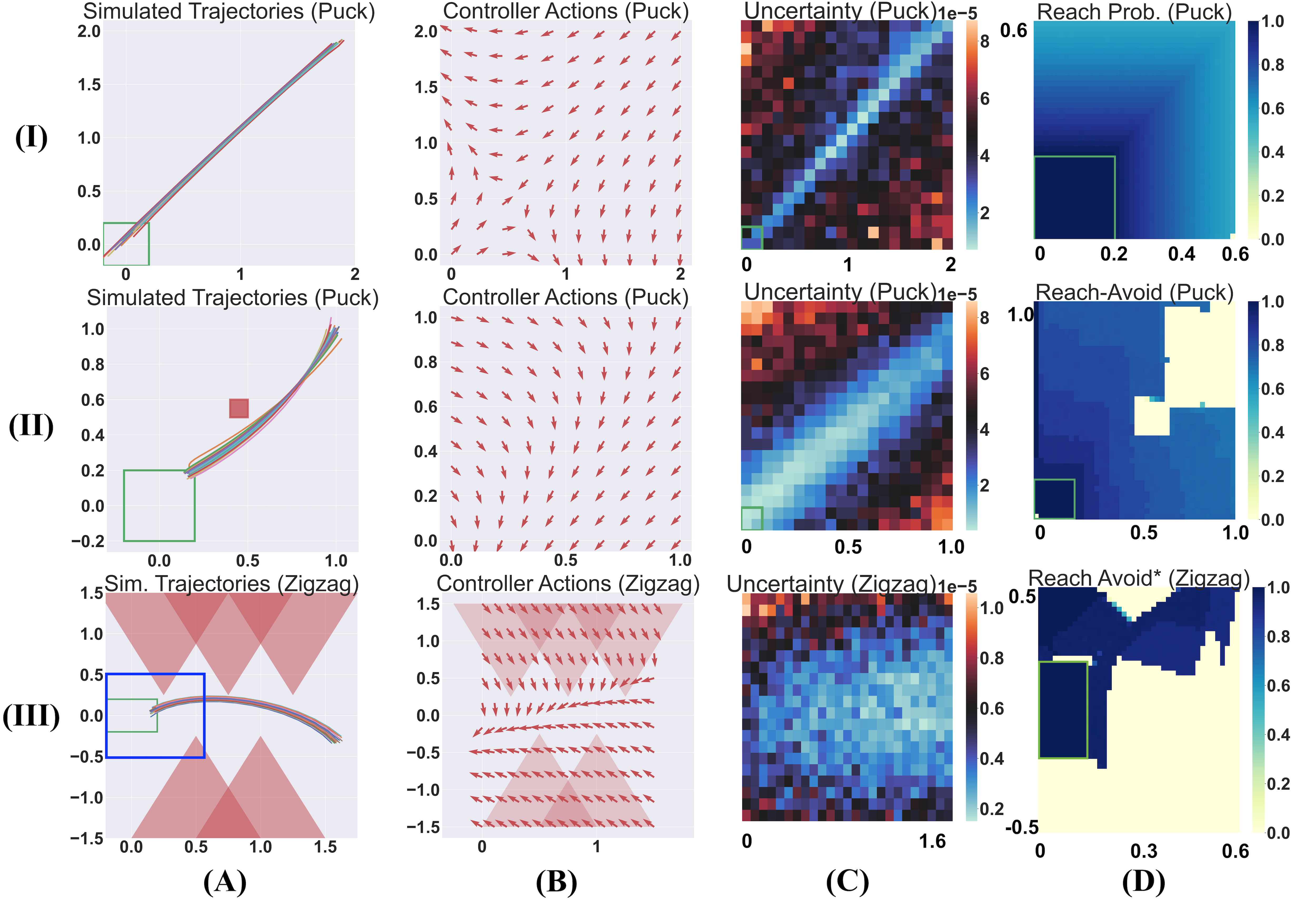}
    \caption{
    \textbf{Top Row (I)}: the Simple Navigation environment with a 2D Puck agent. \textbf{Middle Row (II)}: the obstacle environment with a 2D Puck agent. \textbf{Bottom Row (III)}: the Zigzag environment with 2D Kinematic Car agent. We note that the blue rectangle in column (A) corresponds to area verified in column (D).
    For each environment we analyse the main components of the system.  \textbf{Column (A)}: a collection of 25 simulated runs using the learned policy indicates that the algorithm is successful in learning a policy to reach the goal. \textbf{Column (B)}: the per-point NN control actions show that the controller has learned a reasonable policy even outside of the explored region.  \textbf{Column (C)}: Uncertainty quantification shows that where policy exploration has occurred the BNN is most certain. \textbf{Column (D)}: We are able to verify non-trivial probabilistic guarantees for each system.}
    \label{fig:PlanarEnvs}
\end{figure*}

In this section, we empirically study the effectiveness of our framework on several benchmarks of varying complexity. 
In particular, we consider three different environments (Simple Navigation, Obstacle Avoidance, and Zigzag) and $5$ different agents (2D Kinematic Car \citep{fan2018controller}, 2D Puck \citep{Astrom08}, 3D Hovercraft \citep{Chuchu2020Benchmark}, Ball Robot, and Car Robot \citep{ray2019benchmarking}).
In each setting, we apply Algorithm \ref{alg:backwardsreachability} to certify policies learned via existing model-based strategy synthesis algorithms \citep{chua2018deep}, including an experiment to study the effect of  the parameter choices in Algorithm \ref{alg:backwardsreachability}.
We then proceed to an investigation of our synthesis methodology and an evaluation 
of the tightness of our lower bounds against empirical probability estimates.
Additional details for each of the benchmarks, environments, and agents can be found in the Supplementary Material along with a further discussion of motivation and limitations of our setting. \footnote{Link to code: \textit{github.com/matthewwicker/BNNIterativePrediction}.} 

\paragraph{Experimental Settings}
For BNN and neural policy training we utilize a standard model-based control loop.
Specifically, we learn both model and policy concurrently in an episodic learning framework, whereby we operate in the environment with our policy (following the PE-TS algorithm \citep{chua2018deep}) and aggregate a dynamics dataset on which our BNN model is trained at the end of each episode. The trajectory sampling stage of  PE-TS selects the action sequence which minimizes a cumulative discounted reward. We reward improvement of the agent's distance to the goal with a weighted $l_p$ and, in the presence of obstacles, we add a penalty according to the distance to the obstacles. For all the experiments we initialize our BNN with a Gaussian prior over the parameters; approximate Bayesian inference is performed using Variational Online Gauss-Newton (VOGN) \citep{khan2018fast}. 


\paragraph{Simple Navigation} The first environment we consider is a navigation task where an agent must navigate from any initial state to the origin. Albeit basic, this task becomes challenging with high dimensional agents and noisy sensors. In this scenario we have that the goal region $\mathrm{G}$ 
is a box centered at $(0.05, 0.05)$ (see Figure \ref{fig:PlanarEnvs}(I A)). For the Hovercraft agent the safe set is restricted to be all states with altitude within the interval (0.0, 0.5]. For the Puck, we encode a safe set that restricts the velocity of the object to be less than 1.0 at all times. For the Mujoco agents (Ball and Car Robots) we bound the change in velocity to be less than 0.25 and in these cases (where dimensionality is high) we do not discretize dimensions of the state space which are not safety-critical, e.g., the direction of the main sensor on the Car Robot. In Figure~\ref{fig:PlanarEnvs} row (I), we visualize the actions and simulated trajectories of the Puck agent. We note that the uncertainty of our BNN model is well calibrated, showing higher uncertainty in regions where less data are available. We observe that states with low uncertainty are those for which the lower bound of safely reaching the origin is higher and close to $1$, even order of magnitude time steps away from the goal region (for the experiment we considered $N=30$).

\paragraph{Obstacle Avoidance} The obstacle avoidance task extends the simple navigation environment by adding an obstacle directly between the agent and the goal (see Figure \ref{fig:PlanarEnvs}(II A)). For the Hovercraft, which is not bound to the 2D plane, we assume the obstacle extends infinitely high. 
In Figure~\ref{fig:PlanarEnvs} row (II), we visualize the actions and simulated trajectories of the Puck agent. 
We observe in column (D) that, in this setting, the state-space portion directly behind the obstacle attains a reach-avoid probability of $0$, even though sampled trajectories in Figure \ref{fig:PlanarEnvs}(II A) are able to safely reach the goal region. 
This is due to the conservatism of our approach that computes only a lower bound of $P_{reach}$.

\paragraph{Zigzag} The Zigzag environment is taken from \citep{fan2018controller}. In this task, agents are placed in the fourth quadrant  and are tasked with navigating through a series of equilateral triangles which impede the path to the goal region (see Figure \ref{fig:PlanarEnvs}(III A)). The goal region is a box centered at $(0,0)$. In the Supplementary Material we also analyse a harder version of the Zigzag problem.
In Figure~\ref{fig:PlanarEnvs} row (III), we visualize the actions and simulated trajectories of the Kinematic Car agent. The key observation here is that the large size of the obstacles (see column (D)) makes the verification much more challenging and Algorithm \ref{alg:backwardsreachability} produces overly conservative probabilities for a large portion of initial states.
In what follows (see Table \ref{tab:synth-table}) we will show that,  by modifying the training loss of the agent, as discussed in Section \ref{Sec:Syntesis}, one is able to obtain substantially tighter bounds.

\begin{figure}[h]
    \centering
    \includegraphics[width=0.35\textwidth]{ 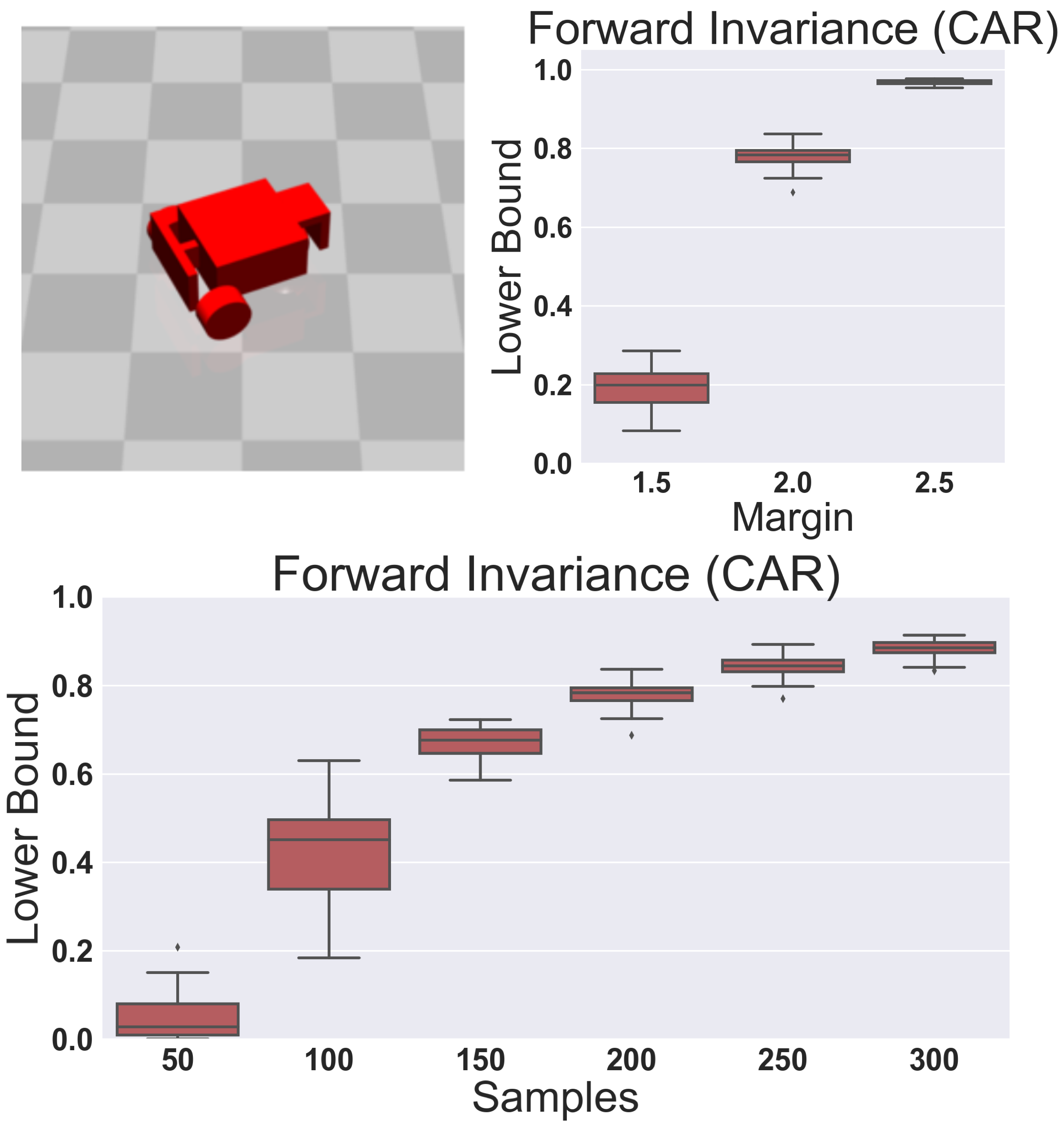}
    \caption{\textbf{Top Left}: Visualization of the car in the Mujoco simulator. \textbf{Top Right}: Increasing the weight margin has a positive effect on the bound. \textbf{Bottom}: Increasing the number of samples considered has a considerably positive effect on the bound. }\label{fig:algparams}
\end{figure}

\paragraph{Effect of Bound Parameters for the Car Robot Agent}
We analyse the effect of Algorithm \ref{alg:backwardsreachability} parameters on a \textit{single-step} prediction on a 25D agent, the Mujoco Car, and the Simple Navigation environment. 
We just focus on a probabilistic version of the forward invariance property considered in \cite{ames2014control}: the policy is considered safe at a given time if the action taken does not move the agent away from the goal region at the next time step.
In Figure~\ref{fig:algparams}, we show how increasing the number of samples from the posterior (parameter $n_s$ in Algorithm \ref{alg:backwardsreachability}), as well as increasing the size of the weight margin (parameter $\rho_{w}$ in Algorithm \ref{alg:backwardsreachability}), improves the resulting lower bound. Intuitively, by increasing the sample size and the weight margin we are able to build a larger under approximation of the projecting weight set. We should stress that, if the weight margin is too large, then this can be detrimental for performance due to increased approximation. 

\paragraph{Synthesis of Certified Strategies}
In Table~\ref{tab:synth-table}, we compare the lower bound obtained from Algorithm \ref{alg:backwardsreachability} with an empirical estimate obtained by simulating $\mathbf{x}_k$ (Eqn \ref{Eqn:SystemEqn}) with a randomly picked initial state (we use $100$ trajectory simulations to compute each empirical estimate). In this case we consider a subset of the verified states in Figure~\ref{fig:PlanarEnvs}, which are close to the goal region and are in the first quadrant.
For each of the tested agents (Puck, 2D Kinematic Car, and Hovercraft) our lower bound is, in the best case, within 0.22 of the empirical estimate and the tightness of the bound is greatly improved when employing Theorem \ref{Th:Syntesis} to synthesise strategies that maximize the lower bound given by Algorithm \ref{alg:backwardsreachability}. In these examples we found that, for the Puck and Hovercraft, synthesised actions allowed us to get a certified safety within 0.03 of the statistically estimated bound. The improvement is expected because our synthesis approach aims to explicitly maximize the lower bound probability and is in line with what was observed for adversarial training of NNs with IBP \citep{gowal2018effectiveness,wicker2021bayesian}.
Further benefits of synthesis can be observed in the Simple Navigation environment, where our approach for strategy synthesis not only improves the certification we provide, but also the empirical performance of the control policy. We further examine this in Figure~\ref{fig:synthvis}, where we observe that with our synthesis algorithm we are able to correct for the erroneous behavior of the original PE-TS controller and certify that virtually all the states have a high probability of reaching the goal.

\begin{table}[]\footnotesize
\begin{tabular*}{0.49\textwidth}{|l|l||l|l||l|l|}
\hline
\textbf{Env.} & \textbf{Agent}      & \textbf{Emp.}  & \textbf{Cert.}  & \textbf{Emp. (S)} & \textbf{Cert. (S)} \\ \hline \hline
Simple & Puck       & 0.738 & 0.4444 & 0.986    & 0.9595    \\ \hline
Zigzag      & 2D Car     & 1.00    & 0.7859 & 1.00       & 0.8550    \\ \hline
Simple  & Hover & 1.00    & 0.6676 & 1.00       & 0.9706    \\ \hline
\end{tabular*}
\caption{\label{tab:synth-table} Lower bound obtained following Algorithm \ref{alg:backwardsreachability} compared to an empirical estimate. \textbf{Emp.} are the empirical estimates each computed over $100$ trajectories simulations. \textbf{Cert.} is the average of the lower bound obtained considering only states in $Q$ where the sampled trajectories start. \textbf{(S)} denotes bounds coming from the control actions synthesized according to our synthesis framework.}
\end{table}

\section{Conclusions}
In this paper we considered iterative predictions with BNNs and studied the problem of computing the probability that a trajectory iteratively sampled from a BNN reaches safely a target goal region. 
We developed methods and algorithms to compute a  lower bound of this reach-avoid probability and synthesize certified neural controllers, based on techniques from dynamic programming and non-convex optimization.
In a set of experiments we show that our framework enables certification of strategies on BNN models and non-trivial, high-dimensional control tasks.

\begin{acknowledgements} 
This project received funding from the ERC under the European Union’s Horizon 2020 research and innovation programme (FUN2MODEL, grant agreement No.~834115).
\end{acknowledgements}

\bibliography{cite}

\onecolumn
\newpage
\title{
Certification of Iterative Predictions in Bayesian Neural Networks: Supplementary Material
}
\maketitle
\appendix

\section{Motivation for BNNs in Control}

Here we provide further details on the motivation to use Bayesian models in a model-based control or reinforcement learning scenario. Choosing an appropriate model when capturing unknown dynamics of a system under consideration is of critical importance to the success of the ultimate algorithm. In particular, it is known that the introduction of slight biases can greatly affect the learning of a good control policy \citep{modelbias97, abbeel2006modelbias}. Model bias can lead to over-confident predictions in the early stages of learning which can in turn lead to unsafe exploration and to the degradation of the learned control policy \citep{abbeel2006modelbias}. Moreover, at deployment time, being able to reason about both out-of-distribution scenarios as well as the uncertainties about ones beliefs regarding the underlying dynamics can enable more safe actions in principle \citep{michelmore2020uncertainty}. This, incorporating a model which is inherently capable of reasoning about uncertainty and which can provide the modeller with critical feedback about model choice is intuitively desirable. 

Bayesian neural networks represent a potentially powerful model for uncertainty-aware model-based reinforcement learning \cite{chua2018deep}. While deterministic neural networks enable greater scalability than Bayesian neural networks, they fail to reason about uncertainty and can be a great source of model bias. Similarly, while GPs tend to be more successful in terms of calibrated uncertainty, they fail to scale to high-dimensional, large-data regime required by many real world problems \citep{pilco}. Bayesian Neural Networks combine the uncertainty benefits of Gaussian processes with the scalability of neural networks. In addition, their uncertainty has been shown to make them more resistant to small changes in their inputs \citep{carbone2020robustness} as well as more sample-efficient during model-based learning \cite{chua2018deep}.

\section{Agent Descriptions and Dynamics}

\paragraph{2D Kinematic Car}

\begin{figure}[h]
    \centering
    \includegraphics[width=0.75\textwidth]{ 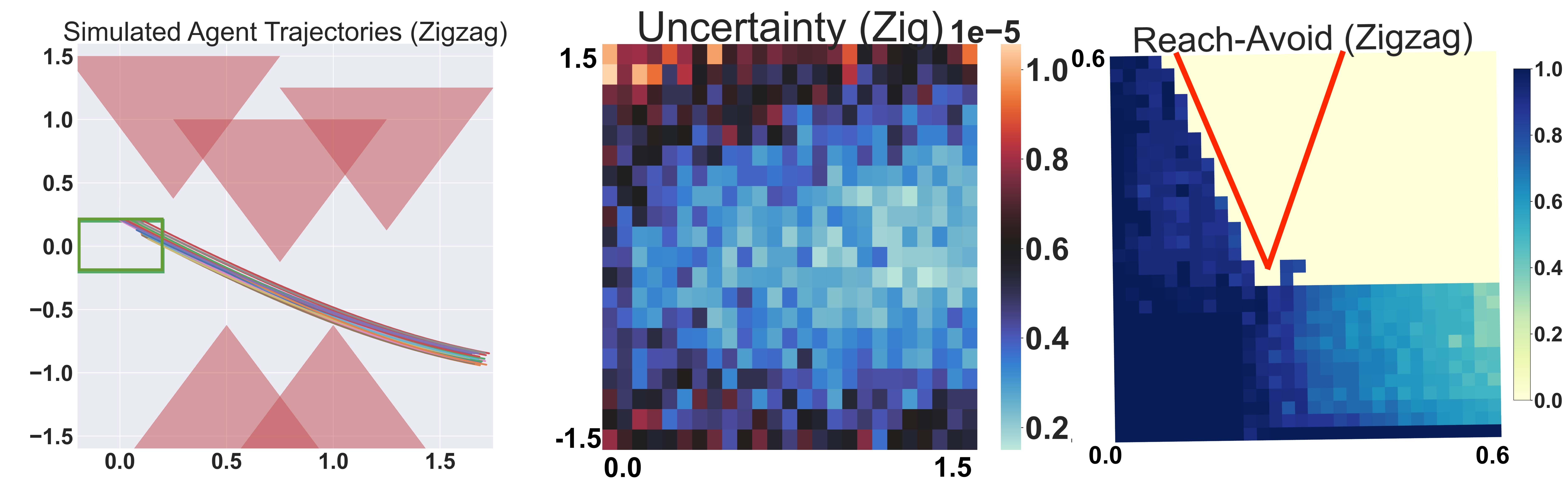}
    \caption{Analysis of the harder variant of Zigzag. \textbf{Left}: Layout of the state-space and 25 simulations from the BNN control loop demonstrates that we learn to solve the problem. \textbf{Center}: The uncertainty lines up well with what is considered in the main text and displays that the model is uncertain in states it is unable to visit. \textbf{Right}: Here we see that unlike what is presented in the main text, the controller has a bias toward navigating the agent upward and thus it is easier to verify the region below the goal. }\label{fig:zighard}
\end{figure}

Both the 2D kinematic car dynamics and the Zigzag environment given in the first row of Figure \ref{fig:PlanarEnvs} are benchmarks from  \citep{fan2018controller}. The agent dynamics model is a planar version of a single rear wheel kinematic vehicle and has three state space variables: two for its position in the plane and one for the rotation status of the wheel. The controller chooses how to change the angle of the wheel as well as the magnitude of its movement vector. We only use the 2D kinematic car in the Zigzag environment where the agent starts below a trench created by a set of five equilateral triangle obstacles. We further consider a `harder' variant of this problem with a more challenging placement of the triangles. For this environment, we use the negative $l_1$ distance from your current position as the reward function, meaning the agent's action is rewarded proportional to its improvement to the goal region. In the standard instance, no information about the obstacles is given and thus must be learned via trial and error at train time. In the harder instance the reward function is modified to compute the $l_2$ distance of the agent to the point of each triangle and the reward is penalized according to this. 

$$
\dot{q} =
  \left[
  \begin{array}{c}
    \dot{x} \\
    \dot{y} \\
    \dot{\theta}\\
  \end{array} 
  \right] = 
  \begin{bmatrix}
    cos(\theta) & 0  \\
    sin(\theta) & 0 \\
    0 & 1  \\
  \end{bmatrix}
  \left[
  \begin{array}{c}
    c_1 \\
    c_2 \\
  \end{array} 
  \right] 
$$

\paragraph{2D Puck}

The Puck environment is derived from a classical control problem of controlling a vehicle from an initial condition to a goal state or way point \citep{Astrom08}. This scenario is slightly more challenging than Zigzag not only due to the increase state-space dimension but also due to the introduction of momentum and reduced control. The state space of the agent is a four vector containing the position in the plane as well as a vector representing the current velocity. The control signal is a two vector representing a change in the velocity (i.e. an acceleration vector). We study this agent in both the `Simple Navigation' and `Obstacle Avoidance' scenarios as visualized in Figure~\ref{fig:PlanarEnvs} rows (I) and (II), respectively. For the former, an $l_2$ reward signal is used, and for the latter an $l_2$ reward penalized by the $l_2$ distance to the obstacle is used.

\begin{figure}[h]
    \centering
    \includegraphics[width=0.45\textwidth]{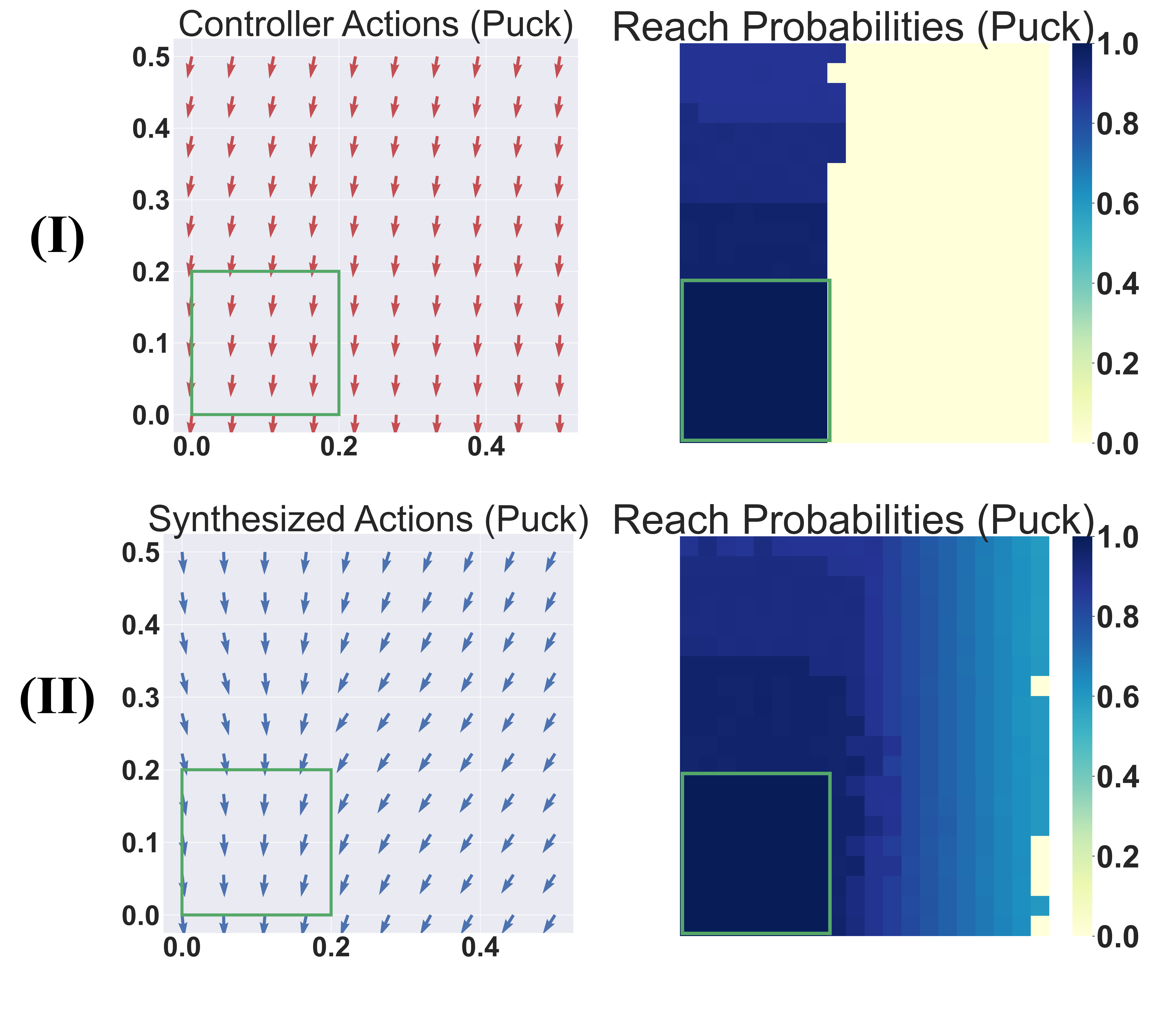}
    \caption{\textbf{Left Column}: Actions given by DNN controller trained with PE-TS (I) do not properly account for movement in the negative $x$ direction compared to synthesized actions (II). \textbf{Right Column}: Verification of learned controller (I) has worse certified bounds compared to that of synthesized control actions (II).}\label{fig:synthvis}
\end{figure}

In Table~\ref{tab:synth-table} we report that our synthesis algorithm not only improves the certification but also the empirical performance of the controller. In Figure~\ref{fig:synthvis}, we give a visual explanation for how this is the case. The original controller learned by the PE-TS algorithm did not, under certain conditions, apply enough velocity in the negative $x$ direction thus not only making certification difficult, but also adversely affecting the performance. We see that our synthesis algorithm corrects for this and enables us to not only improve performance, as reported in Table~\ref{tab:synth-table}, but also improve the certification.

The dynamics of the puck can be given as a the following system of equations where $\eta$ determines friction, $m$ determines the mass of the puck, and $h$ determines the size of the time discretization.

$$
\dot{q} = Aq + Bc 
$$

$$
A =   \begin{bmatrix}
    1 & 0 & h & 0  \\
    0 & 1 & 0 & h  \\
    0 & 0 & 1-h\eta/m & 0  \\
    0 & 0 & 0 & 1-h\eta/m  \\
  \end{bmatrix}
$$

$$
B =   \begin{bmatrix}
    0 & 0  \\
    0 & 0  \\
    h/m & 0  \\
    0 & h/m  \\
  \end{bmatrix}
$$

\paragraph{3D Hovercraft}

The dynamics of the 3D hovercraft are similar to those of the 2D kinematic car, save for the fact that it exists in 3D and must also learn a policy which maintains altitude. This vehicle is taken from \citep{Chuchu2020Benchmark} and amended with a gravity term that makes learning slightly harder, but more realistic. We consider the task of learning to navigate to the goal in the presence of an infinitely tall triangular obstacle placed directly in the path of the agent.  

$$
\dot{q} =
  \left[
  \begin{array}{c}
    \dot{x} \\
    \dot{y} \\
    \dot{z}\\
    \dot{\theta}\\
  \end{array} 
  \right] = 
  \begin{bmatrix}
    cos(\theta) & 0 & 0   \\
    sin(\theta) & 0 & 0\\
    0 & 1 & 0 \\
    0 & 0 & 1 \\
  \end{bmatrix}
  \left[
  \begin{array}{c}
    c_1 \\
    c_2 \\
    c_3 \\
  \end{array} 
  \right] 
$$

\paragraph{Ball Robot}

\begin{figure}[h]
    \centering
    \includegraphics[width=0.75\textwidth]{ 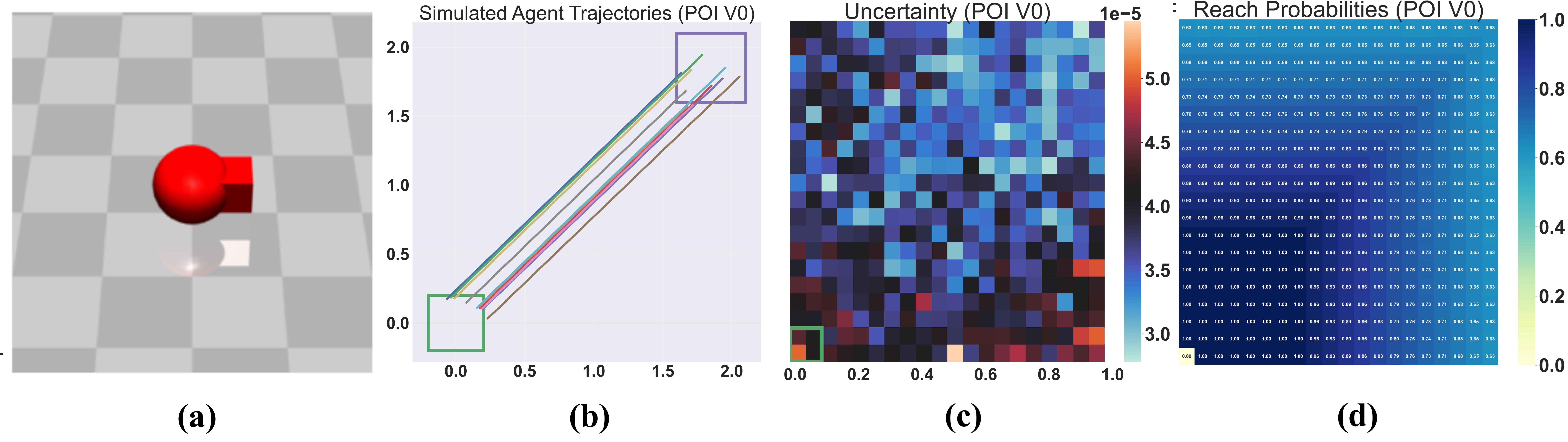}
    \caption{\textbf{Far Left}: A render of the Ball robot in the Mujoco simulator. \textbf{Center Left}: 25 simulations of the Ball roboto control loop. \textbf{Center Right}: The uncertainty of the BNN for the Ball robot does not show an interpretable pattern indicating that it may not be very well calibrated. \textbf{Right}: Despite the higher dimensionality of the Ball robot problem we are still able to compute good lower-bounds for a portion of the state-space.}\label{fig:poi_supplement}
\end{figure}

For the 3D Ball Robot environment we consider a simple locomotive task inside of the OpenAI gym RL suite \citep{brockman2016openai}. The agent observes a set of noisy sensor outputs involving the center of mass of the robot, the status of its wheel, velocity, and the location of its sensor (red cube seen in [TODO]). The challenge of this task is not only the dimensional of the robots state-space but also noise injected into the observations. Due to this, the BNN dynamics model takes longer to converge and it makes it challenging for the PE-TS algorithm to quickly identify a strong strategy. Despite this, after 20 episodes of learning, the BNN fits the dynamics of the system well and is able to reliably navigate to the goal. The dynamics of this agent are wholly defined by the Mujoco physics simulator and thus are too complex to be listed in closed form here.

\paragraph{Car Robot}

\begin{figure}[h]
    \centering
    \includegraphics[width=0.75\textwidth]{ 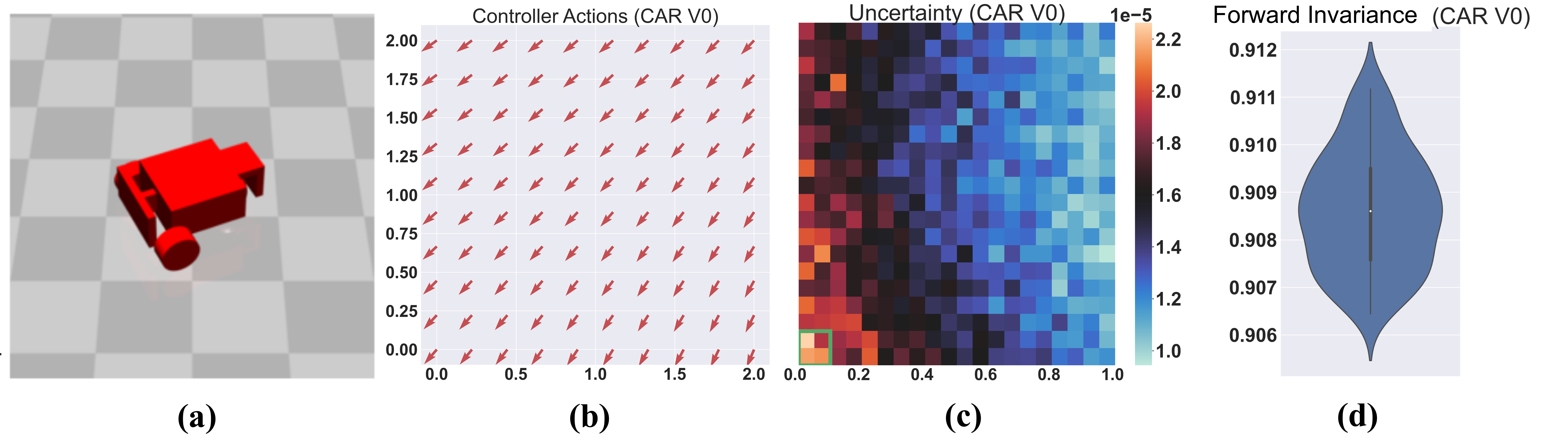}
    \caption{\textbf{Far Left}: A render of the Car robot in the Mujoco simulator. \textbf{Center Left}: A visualization of the control policy learned by the car robot. \textbf{Center Right}: The uncertainty of the BNN for the Car robot does not show an interpretable pattern indicating that it may not be very well calibrated. \textbf{Right}: Similarly to what is shown in Figure~\ref{fig:algparams} we are able to obtain high probability certificates for the one-step forward invariance property.}\label{fig:car_supplement}
\end{figure}

The Car Robot poses a locomotion problem in which the controller must navigate a vehicle with two independently-driven parallel wheels and a free rolling rear wheel into a goal region. Just as with the ball robot, the Car is not fixed to the 2D-plane, and observations are taken from noisy sensors about the state of the robot in space. This problem is significantly more challenging than the previous robots as moving forward and turning require coordination of the independently actuated wheels. Due to the large dimensionality of the state space of this robot, reach properties cannot be easily handled due to the explosion of the state space discretization needed. Instead, we compute a probabilistic lower bound on the one step forward invariance property presented in \citep{ames2014control}. Similar to the Ball robot, the dynamics of the physics simulator are too complex to be listed in closed form here.

\begin{figure}[h]
    \centering
    \includegraphics[width=0.85\textwidth]{ 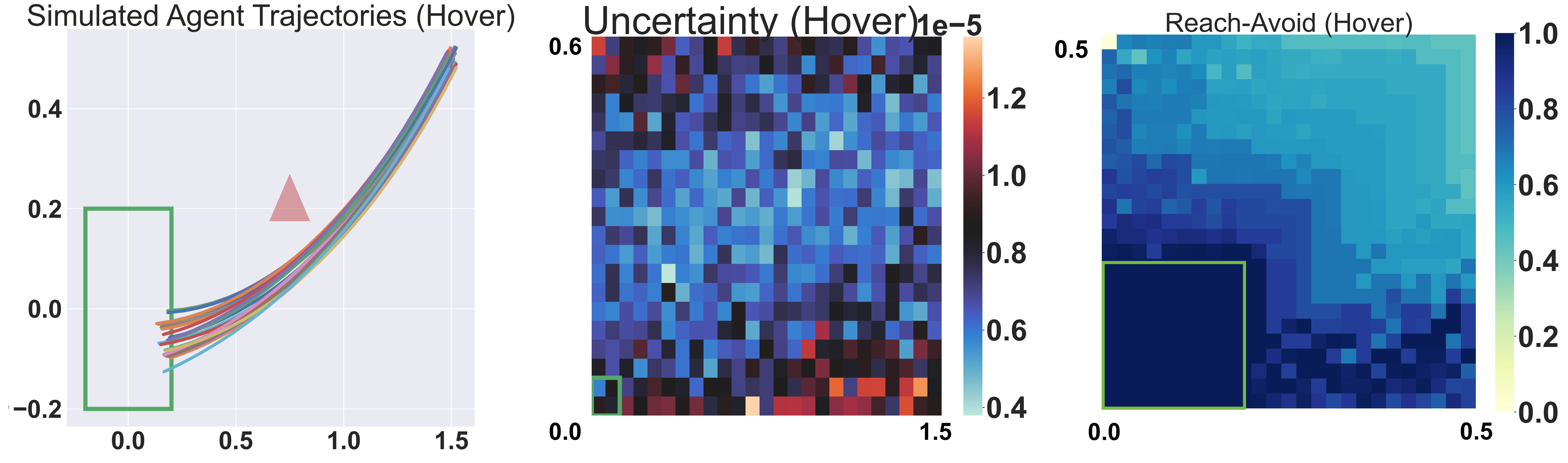}
    \caption{\textbf{Left:} Simulated trajectories from the hovercraft control loop demonstrates that it successfully solves the task well and navigates around the obstacle. \textbf{Center}: The uncertainty of this BNN is interpretable but may not be considered well-calibrated. \textbf{Right}: The reach-avoid property shows that we are able to verify a large portion of the state-space, here we are considering negative altitude as the avoid region of the state-space.}\label{fig:hov_supplement}
\end{figure}

\section{Learning Parameters}

In this section, we use Table~\ref{tab:params} to report the parameters of our  learning set up. We report the state-space and control dimensions ($n$ and $m$ respectively) as well as the architectures used for both the dynamics model (BNN) and controller (DNN). We give the number of episodes of PE-TS that are used in these scenarios as well as the number of samples and time horizon considered during the trajectory sampling stage of PE-TS. During the trajectory sampling stage we have a set of 3-10 different discrete control signals which can be picked for each control dimension (see dynamics above). In order to pick the best action for the current time step, one simple randomly samples a series of actions over a finite time horizon and uses the sum of discounted rewards in order to pick the most promising action sequence. From this, only the first action is taken and then the process is repeated. For more details see \cite{chua2018deep} and for the exact sampling parameters for our settings we reference Table~\ref{tab:params}.

All BNN posteriors are approximately inferred with the VOGN algorithm \cite{khan2018fast}.

\begin{table}[h]\footnotesize
\begin{tabular*}{1.0\textwidth}{|l|l|l|l|l|l|l|l|}
\hline
Benchmark:   & n, m & BNN Arch. & Control Arch. & Property           & Episodes & Action Samples & Time Horizon \\ \hline
Zigzag       & 3,2  & FCN-1-64  & FCN-1-32      & Reach-Avoid        & 8        	&           4096          &      5                                   \\
Zigzag-Hard     & 3,2  & FCN-1-64  & FCN-1-64      & Reach-Avoid        & 12       &      8000               &    3                                     \\
PointMass (Simple) & 4,2  & FCN-1-64  & FCN-1-64      & Reach              & 10       	&       2048              &   9                                      \\
PointMass (Obstacle) & 4,2  & FCN-1-64  & FCN-1-32      & Reach-Avoid        & 10     	  &        7500             &    10                                     \\
Hovercraft   & 5,3  & FCN 1-64  & FCN 1-32      & Reach-Avoid        & 15       	&         10000            &         5                                \\
Ball Robot   & 8,2  & FCN-1-128 & FCN-1-64      & Reach              & 20       		&           10000          &          10                               \\
Car Robot    & 23,2 & FCN-1-128 & FCN-1-64      & Forward Invariance & 25       &           5000          &           5                              \\ \hline
\end{tabular*}    \caption{Here we report the parameters of our learning environments. Episodes referes to the number of learning episodes performed before the policy was successful at solving the task. Action Samples and Time Horizon refer to the trajectory sampling required by PE-TS. }\label{tab:params}
\end{table}

\section{Limitations of the Approach}

The limitations of our approach come from a few places: approximation of Bayesian inference, approximation in certification, and the reliance on discretization of the state space. 

\paragraph{Approximate Inference} In both theory and practice, an overly approximate inference method can hamper our framework. While it makes perfect sense to verify with respect to ones posterior beliefs about an underlying system’s unknown dynamics, if those posterior beliefs are flawed (e.g. due to overly approximate inference) then the probabilities of safety may not match the true probability of safety due to a gap between posterior belief and reality. We do note, however, that recent works in approximate inference have shown that with a larger computational budget that BNNs inferred (even with MCMC) can reliably scale to inputs with tens of thousands of dimensions. 

\paragraph{Reliance on Discretization} While modern certification and bound propagation techniques for NNs and BNNs have shown remarkable scalability, being able to scale to images with thousands of input dimensions, we remark that formal reach-avoid certification of an input space can only be done by fully considering the potential interdependence of different dimensions of the state space. As such, while single-step certification may be able to scale to large inputs, our method incurs a complexity exponential in the number of state space variables that one discretizes over. Finally we note that, as in our experiments, one can chose to only discretize safety critical variables while also providing certification for single-step constraints over other variables.

\section{Proofs}

\textbf{Proof of Proposition \ref{Prob:verification}}
In what follows, we omit $\pi$ (which is given and held constant) from the probabilities for a more compact notation. 
The proof is by induction. The base case is $k=N$, for which we have
$$V_N^{\pi}(x) =\mathbf{1}_\mathrm{G}(x)= P_{reach}(\mathrm{G},\mathrm{S},x,[N,N]),  $$
which holds trivially. Under the assumption that, for any given $k\in [0,N-1]$, it holds that 
\begin{align}
    \label{Eqn:IndBaseCase}
  V_{k+1}^{\pi}(x) = P_{reach}(\mathrm{G},\mathrm{S},x,[k+1,N]),
\end{align}
we show the induction step for time step $k$. 
In particular, 
\begin{align*}
     &P_{reach}(\mathrm{G},\mathrm{S},x,[k,N]|\pi)=\\
     &Pr( \mathbf{x}_k \in \mathrm{G}  |\mathbf{x}_k = x) +\sum_{j=k+1}^{N}Pr( \mathbf{x}_j \in \mathrm{G} \wedge \forall j' \in [k,j) , \mathbf{x}_{j'} \in \mathrm{S} |\mathbf{x}_k= x)= \\
     &\mathbf{1}_{\mathrm{G}}(x) + \mathbf{1}_{\mathrm{S}}(x)\sum_{j=k+1}^{N}Pr( \mathbf{x}_j \in \mathrm{G} \wedge \forall j' \in [k,j), \mathbf{x}_{j'} \in \mathrm{S} |\mathbf{x}_k= x)
\end{align*}     
Now in order to conclude the proof we want to show that
\begin{align*}
    \sum_{j=k+1}^{N}Pr( \mathbf{x}_j \in \mathrm{G} \wedge \forall j' \in [k,j)+1, \mathbf{x}_{j'} \in \mathrm{S} |&\mathbf{x}_k= x) =\\
    & \int V_{k+1}^{\pi}(\bar{x})p(\bar{x} \mid (x,\pi_k(x)),\mathcal{D} ) d\bar{x}.
\end{align*}  
This can be done as follow
\begin{align*}
   & \sum_{j=k+1}^{N}Pr( \mathbf{x}_j \in \mathrm{G} \wedge \forall j' \in [k+1,j), \mathbf{x}_{j'} \in \mathrm{S} |\mathbf{x}_k= x)=\\
 & Pr( \mathbf{x}_{k+1} \in \mathrm{G} |\mathbf{x}_k=x)+ \\
 &\quad \sum_{j=k+2}^{N}Pr( \mathbf{x}_j \in \mathrm{G} \wedge \forall j' \in [k+1,j), \mathbf{x}_{j'} \in \mathrm{S} |  \mathbf{x}_k= x)=\\
   & \int_{\mathrm{G}} p(\bar{x} \mid (x,\pi_k(x)),\mathcal{D} )d\bar{x} +\\
   &\quad \sum_{j=k+2}^{N}\int_{\mathrm{S}}Pr( \mathbf{x}_j \in \mathrm{G} \wedge \forall j' \in [k+2,j), \mathbf{x}_{j'} \in \mathrm{S} \wedge \mathbf{x}_{k+1}=\bar{x} |  \mathbf{x}_k= x) d\bar{x}=\\
  & \int_{\mathrm{G}} p(\bar{x} \mid (x,\pi_k(x)),\mathcal{D} )d\bar{x} +\\
  & \quad \sum_{j=k+2}^{N}\int_{\mathrm{S}} Pr( \mathbf{x}_j \in \mathrm{G} \wedge \forall j' \in [k+2,j), \mathbf{x}_{j'} \in \mathrm{S}   | \mathbf{x}_{k+1}=\bar{x} )p(\bar{x} \mid (x,\pi_k(x)),\mathcal{D} )d{\bar{x}}=\\
   & \int \big( \mathbf{1}_{\mathrm{G}}(\bar{x}) +\\
   &\quad \mathbf{1}_{\mathrm{S}}(\bar{x}) \sum_{j=k+2}^{N} Pr( \mathbf{x}_j \in \mathrm{G} \wedge \forall j' \in [k+2,j), \mathbf{x}_{j'} \in \mathrm{S}   | \mathbf{x}_{k+1}=\bar{x} )\big)p(\bar{x} \mid (x,\pi_k(x)),\mathcal{D} )d{\bar{x}}=\\
    & \int V_{k+1}^{\pi}(\bar{x})p(\bar{x} \mid (x,\pi_k(x)),\mathcal{D} )d{\bar{x}}\\
\end{align*}
where the third step holds by application of Bayes rule over multiple events.

\textbf{Proof of Theorem \ref{th:VerificationLoerBound}}
The proof is by induction. The base case is $k=N$, for which we have
$$\inf_{x\in q}V_N^{\pi}(x)=\inf_{x\in q}\mathbf{1}_{\mathrm{G}}(x)=\mathbf{1}_{\mathrm{G}}(q)=K_N^{\pi}(q). $$
Next, under the assumption that for any $k\in \{0,N-1 \}$ it holds that 
\begin{align*}
 \inf_{x\in q} V_{k+1}^{\pi}(x) \geq K_{k+1}^{\pi}(q),
\end{align*}
we can work on the induction step: 
in order to derive it, it is enough to show that for any $\epsilon >0$
\begin{align*} 
&\int  V_{k+1}^{\pi}(\bar{x})p(\bar{x} \mid (x,\pi_k(x)),\mathcal{D} )d\bar{x} \geq \\ &\quad F([-\epsilon,\epsilon]|\sigma^2)^n \sum_{i=1}^{n_p}\int_{H^{q,\pi}_{k,i}}  v_{i-1}  p_{\mathbf{w}}(w|\mathcal{D})  dw,
\end{align*}
where $F([-\epsilon,\epsilon]|\sigma^2)=\text{erf}(\frac{\epsilon}{\sqrt{2 \sigma^2}})$ is the cumulative function distribution for a normal random variable with zero mean and variance $\sigma^2$ being within $[-\epsilon,\epsilon].$
This can be argued by rewriting the first term in parameter space (recall that the stochastic kernel $T$ is induced by $p_{\mathbf{w}}(w|\mathcal{D})$) and providing a lower bound, as follows: 
\begin{align*}
    & \int  V_{k+1}^{\pi}(\bar{x})p(\bar{x} \mid (x,\pi_k(x)),\mathcal{D} )d\bar{x}= \\ 
    & \text{(By definition of predictive distribution)}  \\
    & \int \big( \int  V_{k+1}^{\pi}(\bar{x}) p(\bar{x}|(x,u),w) d\bar{x} \big) p_{\mathbf{w}}(w|\mathcal{D})  dw \geq \\ 
     & \text{(By $V_{k+1}^k$ being non negative everywhere and by the Gaussian likelihood)}  \\
     & \int \big( \int_{f^{w}(x,\pi(x,k))+\epsilon}^{f^{w}(x,\pi(x,k)-\epsilon}  V_{k+1}^{\pi}(\bar{x})\mathcal{N}(\bar{x}|f^{w}(x,\pi(x,k)),\sigma^2\cdot I) d\bar{x} \big) p_{\mathbf{w}}(w|\mathcal{D})  dw\geq \\ 
       & \text{(By standard inequalities of integrals)}  \\
  & \int  \inf_{\bar{\gamma}\in [-\epsilon,\epsilon]}  V_{k+1}^{\pi}(f^{w}(x,\pi(x,k)+\bar{\gamma}) \big( \int_{[-\epsilon,\epsilon]^n} \mathcal{N}(\gamma|0,\sigma^2) d\gamma \big)^n p_{\mathbf{w}}(w|\mathcal{D})  dw\geq \\  
       & \text{(By the assumptions that for $i\neq j$ $H^{q,\pi}_{k,i}$ and $H^{q,\pi}_{k,j}$ are non-overlapping)}  \\
    &\big( \int_{[-\epsilon,\epsilon]} \mathcal{N}(\gamma|0,\sigma^2) d\gamma \big)^n \sum_{i=1}^{n_p}\int_{H^{q,\pi,\epsilon}_{k,i}}  \inf_{\bar{\gamma}\in [-\epsilon,\epsilon]}  V_{k+1}^{\pi}(f^{w}(x,\pi(x,k)+\bar{\gamma})   p_{\mathbf{w}}(w|\mathcal{D})  dw,\\
       & \text{(By the fact that $v_i\leq \inf_{x\in q} V_{k+1}^{\pi}(f^{w}(x,\pi(x,k)+\bar{\gamma})$ )}  \\
    &\big( \int_{[-\epsilon,\epsilon]} \mathcal{N}(\gamma|0,\sigma^2) d\gamma \big)^n \sum_{i=1}^{n_p}v_i \int_{H^{q,\pi,\epsilon}_{k,i}}   p_{\mathbf{w}}(w|\mathcal{D})  dw,
\end{align*}
     where the last step concludes the proof  because, by the induction hypothesis, we know that for $q' \subseteq \mathbb{R}^n$
     $$
\inf_{\bar{x}\in q'}     V_{k+1}^{\pi}(\bar{x})\geq K_{k+1}^\pi(q')
     $$ 
     and by the construction of sets $H^{q,\pi}_{k,i}$ for each of its weights  $K_{k+1}^\pi(f^w(\bar{x},\pi(x,k))$ is lower bounded by $v_{i-1}$.

\end{document}